\documentclass[conference]{IEEEtran}
\IEEEoverridecommandlockouts
\usepackage{cite}

\usepackage{amsmath,amssymb,amsfonts}
\usepackage{algorithmic}
\usepackage{graphicx}
\usepackage{textcomp}
\usepackage{xcolor}
\usepackage[ruled,vlined]{algorithm2e}
\usepackage{subfig}
\usepackage{caption}
\usepackage{comment}
\usepackage{hyperref}
\hypersetup{
    colorlinks=true,
    linkcolor=blue,
    filecolor=magenta,      
    urlcolor=cyan,
    pdftitle={Overleaf Example},
    pdfpagemode=FullScreen,
    }
\def\BibTeX{{\rm B\kern-.05em{\sc i\kern-.025em b}\kern-.08em
    T\kern-.1667em\lower.7ex\hbox{E}\kern-.125emX}}
\begin{document}

\title{Human Decision Makings on Curriculum Reinforcement Learning with Difficulty Adjustment}

\author{
Yilei Zeng*, Jiali Duan*, Yang Li
Emilio Ferrara, Lerrel Pinto, C.-C. Jay Kuo, Stefanos Nikolaidis\\
\{yilei.zeng,jialidua,yli546,emiliofe,nikolaid\}@usc.edu \\
lerrel@cs.nyu.edu,\quad cckuo@sipi.usc.edu \\
}

\maketitle

\begin{abstract}
Human-centered AI considers human experiences with AI performance. While abundant research has been helping AI achieve superhuman performance either by fully automatic or weak supervision learning, fewer endeavors are experimenting with how AI can tailor to humans' preferred skill level given fine-grained input. In this work, we guide the curriculum reinforcement learning results towards a preferred performance level that is neither too hard nor too easy via learning from the human decision process. To achieve this, we developed a portable, interactive platform that enables the user to interact with agents online via manipulating the task difficulty, observing performance, and providing curriculum feedback. Our system is highly parallelizable, making it possible for a human to train large-scale reinforcement learning applications that require millions of samples without a server. The result demonstrates the effectiveness of an interactive curriculum for reinforcement learning involving human-in-the-loop. It shows reinforcement learning performance can successfully adjust in sync with the human desired difficulty level. We believe this research will open new doors for achieving flow and personalized adaptive difficulties. Our demo executable and videos are available at \url{https://bit.ly/372vCNv}.

\end{abstract}

\section{Introduction}
Humans make billions of decisions in games, and how to leverage this wealth of resources to make better adaptive and personalized systems has been a perpetual pursuit. Humans are both quick and impatient learners, as they lose interest when they outgrow once challenging games or stagnate too early. By learning human's learning process through gaming feedback loops, AI can better create a flow channel that is neither too challenging nor too boring. A curriculum organizes the learning process in an upward spiral by gradually mastering more complex skills and knowledge~\cite{bengio2009curriculum}. When combined with reinforcement learning, it's been shown that a curriculum can improve convergence or performance compared to learning from the target task from scratch~\cite{taylor2009transfer,graves2017automated,florensa2017reverse}. Thus, will make finer adjustments faster.

\begin{figure}[!thb]
      \centering
      \includegraphics[width=0.48\textwidth]{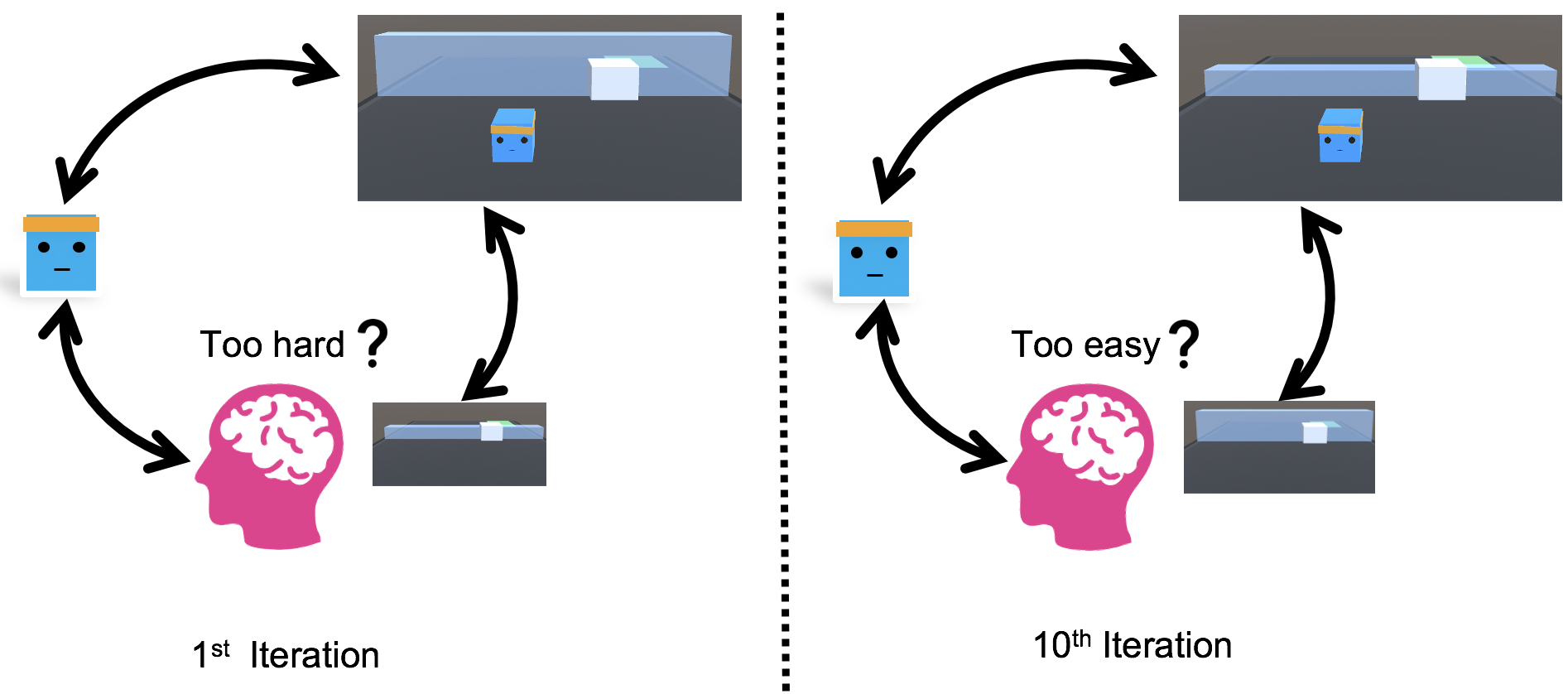}
      \caption{Given specific scenarios during curriculum training, humans can adaptively decide whether to be ``friendly'' or ``adversarial'' by observing the progress the agent is able to make. In cases where performance degrades, a user may
      flexibly adjust the strategy as opposed to an automatic assistive agent.}
      \label{fig:env-signal}
      \vspace{-1em}
\end{figure}

Previous works~\cite{bengio2009curriculum,held2018automatic} focus on reaping the advantage of a curriculum strategy to train the best performing AI agent, via automatically proposing curriculum through another RL agent, such as teacher-student framework~\cite{matiisen2019teacher,portelas2019teacher}, self-play~\cite{sukhbaatar2017intrinsic,bansal2017emergent,baker2019emergent} or goal-gan~\cite{held2018automatic}. One way of interpreting these approaches is that curriculum evolves through the adversarial nature between the two agents, similar to GAN~\cite{goodfellow2014generative,duan2018portraitgan}.

Compared to an automatic agent, human has an innate ability to improvise and adapt when confronted with different scenarios, in order to design more personalized experiences we must capture these human indications, to help the expalinability and flexibility for curriculum reinforcement learning. In Figure~\ref{fig:env-signal}, a user is able to intuitively understand the learning progress and dynamically manipulate the task difficulty by changing the height of the wall. With new challenging environments, we show how human inductive bias can help solve three nontrivial tasks to various difficulty levels that are otherwise unsolvable by learning from scratch or even auto-curriculum.

Another key motivation is assistive agents such as autonomous driving systems, language-based virtual systems, and robotic companions. The agent should provide services adjusted to human preferences and personal needs~\cite{zeng2021human}.

In Section~\ref{sec:related}, we give a brief introduction of related work. In Section~\ref{sec:platform}, Our interactive curriculum platform is introduced, with which we identify the ``inertial'' problem in an ``easy-to-hard'' automatic curriculum. In Section~\ref{sec:experiment}, We show preliminary results of user studies on our environments that require millions of interactions. We conclude and discuss future work in Section~\ref{sec:conclusion}. Finally, we outline some limitations for this line of research in Section~\ref{sec:limitations}.

\section{Related Work}\label{sec:related}
\subsection{Curriculum Reinforcement Learning}
 Apart from previously mentioned automatic learning methods, the most related work to ours is~\cite{heess2017emergence}, which shows empirically how a rich environment can help to promote the learning of complex behavior without explicit reward guidance. In comparison, we evolve environments leveraging human's inductive bias in curriculum design.

\subsection{Human-in-the-Loop Reinforcement Learning}
As learning agents move from research labs to the real world, it becomes increasingly important for human users, especially those without programming skills, to teach agents desired behavior. A large amount of work focuses on imitation learning~\cite{schaal1999imitation,ross2011reduction,ho2016generative,pinto2016supersizing}, where demonstrations from the expert act as direct supervision. Humans can also interactively shape training with only positive or negative reward signals~\cite{knox2009interactively} or combine manual feedback with rewards from MDP~\cite{knox2010combining,abel2017agent}. A recent work formulates human-robot interaction as an adversarial game~\cite{duan2019robot} and shows improvement in grasping success and robustness when the robot trains with a human adversary. 

In this paper, we aim to close the loop between these two fields by studying the effect of interactive curriculum on reinforcement learning. To achieve this, we have designed three challenging environments that are nontrivial to solve even for state-of-the-art RL method~\cite{schulman2017proximal}, which we describe in the next section. 

\section{Interactive Curriculum Guided by Human} \label{sec:platform}
\subsection{Interactive Platform}
\label{sec:interactive-platform}

We build our interactive platform with three goals in mind:
1) Real-time online interaction with flexibility;
2) Parallelizable for human-in-the-loop training
3) Seamless control between reinforcement learning and human-guided curriculum. 
\begin{figure}[h]
      \centering
      \includegraphics[width=0.48\textwidth]{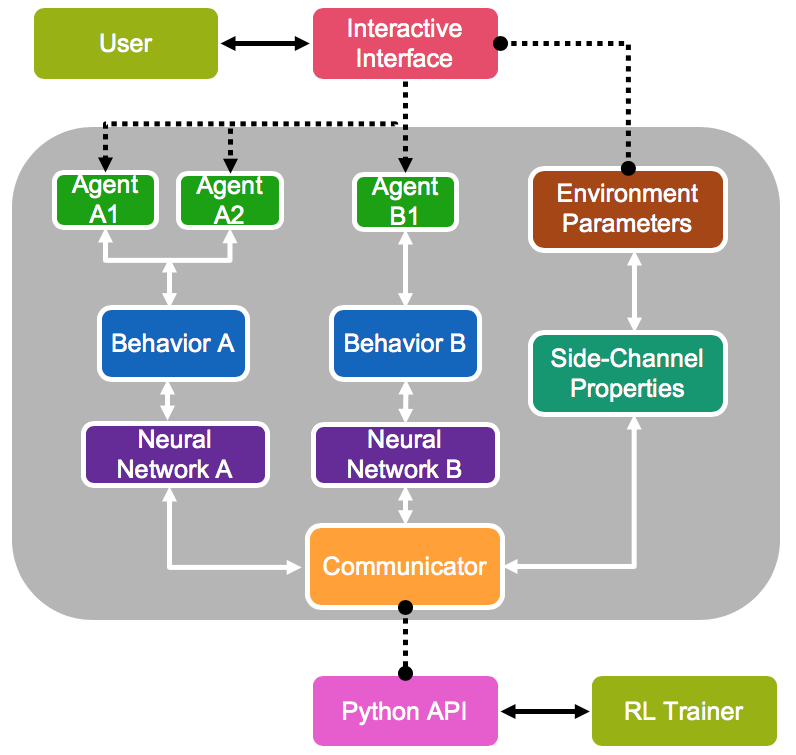}
      \caption{General design of our interactive platform and associations between environment container with RL trainer as well as interactive-interface.}
      \label{fig:interactive-env}
\end{figure}

\begin{figure}[h]
      \centering
      \includegraphics[width=0.48\textwidth]{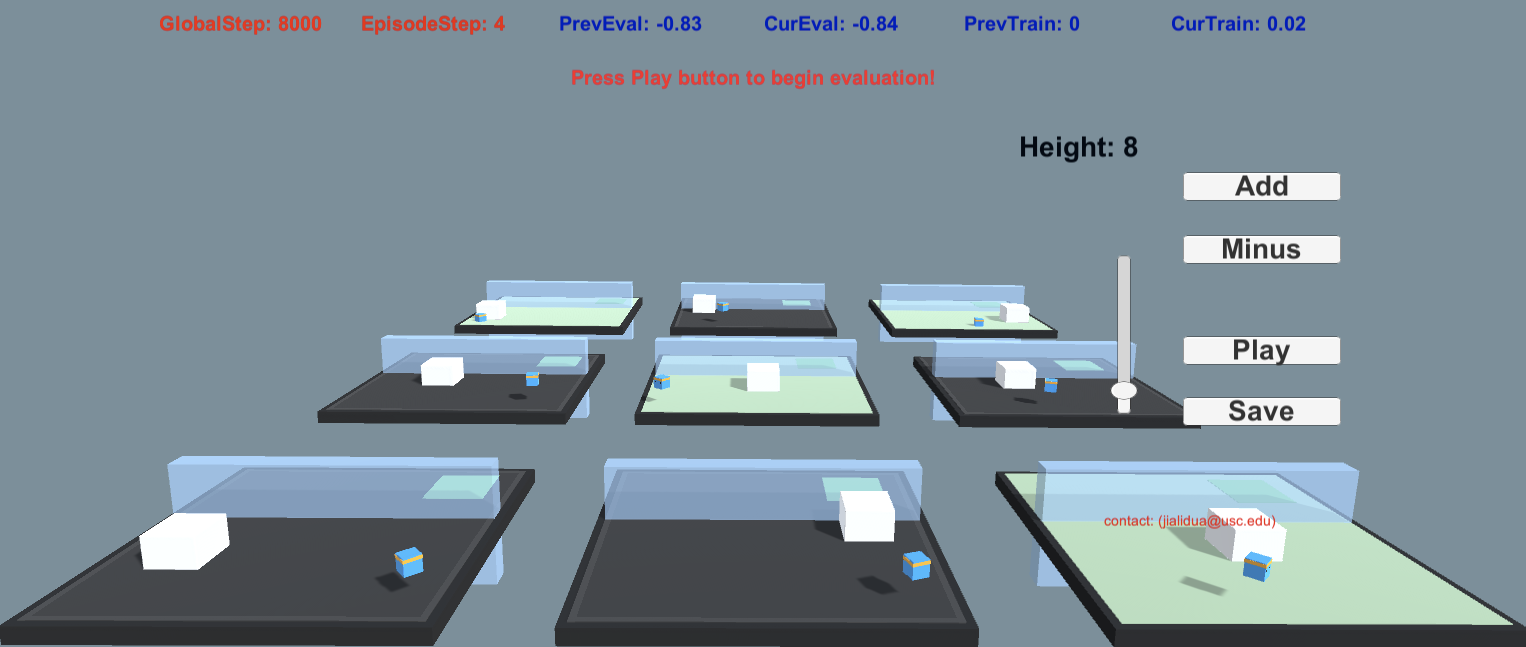}
      \caption{Example of our interactive platform training in parallel.}
      \label{fig:parallel-env}
\end{figure}

We run an event-driven environment container separated from the training process to achieve the first goal, allowing the user to send a control signal (e.g., UI control, scene layout, task difficulty) to the environment during training via the interactive interface. The framework is shown in Figure~\ref{fig:interactive-env} to explain the positions users, environment and training algorithms stand. We integrate human-interactive signals into RL parallelization to achieve similar efficiency as automatic training. An example for parallel training is shown in Figure~\ref{fig:parallel-env}. We perform centralized SGD updates with decentralized experience collection as agents of the same kind share the same network policy~\cite{mnih2016asynchronous}. We also enable controlling environment parameters in different instantiations simultaneously via a unified interactive interface, making it possible to solve tasks requiring millions of interactions. For the third goal, we display real-time instructions and allow users to inspect learning progress when designing the curriculum.

\begin{figure*}[!thb]
      \centering
      \includegraphics[width=0.95\textwidth]{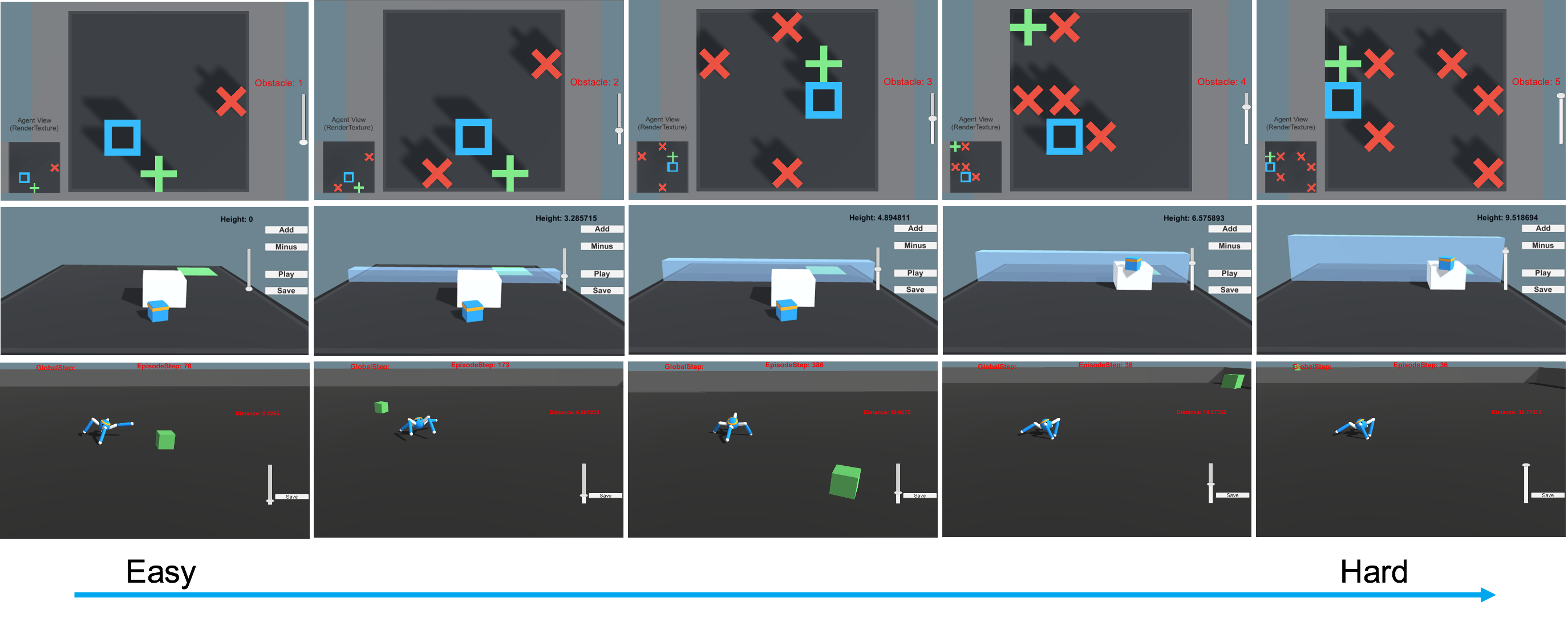}
      \caption{Our interactive platform for curriculum reinforcement learning allows the user to manipulate the task difficulty via a unified interface (slider and buttons). All three tasks receive only sparse rewards. The manipulable variable for the three environments is respectively the number of red obstacles (GridWorld, Top row), the height of the wall (Wall-Jumper, Middle row), and the radius of the target (SparseCrawler, Bottom row). The task difficulty gradually increases from left to right.}
      \label{fig:curriulum-env}
\end{figure*}
Figure~\ref{fig:curriulum-env} shows our released environments for curriculum reinforcement learning, where users can manipulate the task difficulty. The agents will reach the green target in GridWorld, navigate to land on the green mat in Wall-Jumper and reach the dynamic green box in SparseCrawler, respectively. As shown in Figure~\ref{fig:curriulum-env}, the user formulated curriculum in a way that was neither too hard nor too easy for the agent, so as to maximize the efficiency and quality trade-off. During interaction, the user can pause, play or save the current configuration. The locations of the objects in the arena are customizable with cursor and the height of the wall is tunable for difficulty transitions. Our interactive interface are same for the rest environments listed below. \\
\textbf{Grid-World} The agent (represented as a blue square) is tasked to reach the goal position (green plus), by navigating through obstacles (red cross, maximally 5). All objects are randomly spawn on a 2D plane. A positive reward 1 for reaching the goal, negative 1 for cross and -0.01 for each step. Movements are in cardinal directions.\\
\textbf{Wall-Jumper} The goal is to navigate a wall (maximum height is 8), by jumping or (possibly) leveraging a block (white box). Positive reward of 1 for successful landing on the goal location (green mat) or negative 1 for falling outside or reaching maximum allowed time. A penalty of -0.0005 for each step taken. The observation space is 74 dimensional, corresponding to 14 ray casts each detecting 4 possible objects, plus the global position of the agent and whether or not the agent is grounded. Allowed actions include translation, rotation and jumping.\\
\textbf{Sparse-Crawler} A crawler is an agent with 4 arms and 4 forearms. The aim is to reach a randomly located target on the ground (maximum radius of 40). The state is a vector of 117 variables corresponding to position, rotation, velocity, and angular velocities of each limb plus the acceleration and angular acceleration of the body. Actions space is of size 20, corresponding to target rotations for joints. Only sparse reward is provided, when the target is reached.

\begin{figure*}[!thb]
\centering
    \subfloat[Training curve]{\label{fig:inertial-train}
    \includegraphics[width = 0.70\columnwidth]{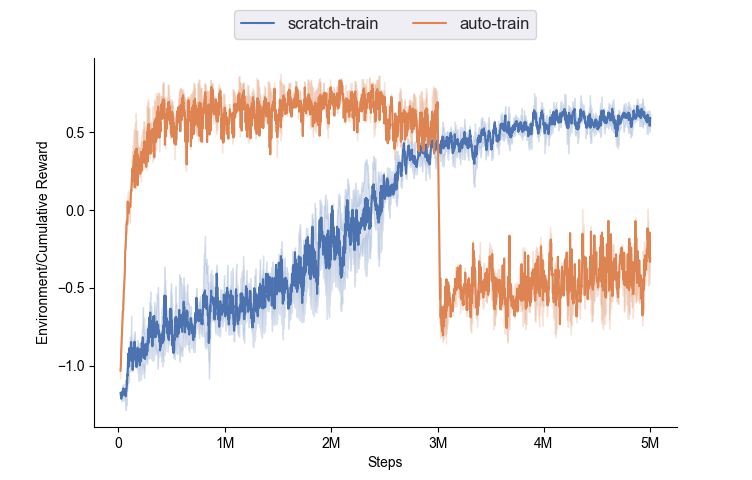}}
   \subfloat[Testing curve]{\label{fig:inertial-test}
   \includegraphics[width = 0.70\columnwidth]{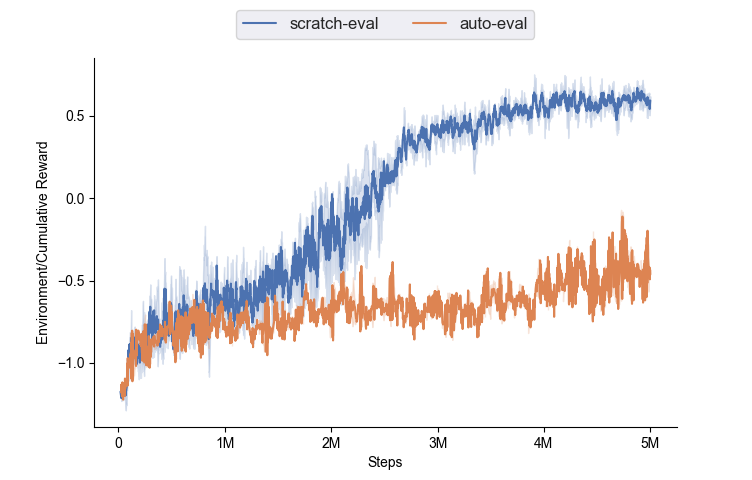}}
   \subfloat[High Wall]{\label{fig:inertial-wall}
   \raisebox{1.2em}{\includegraphics[width = 0.65\columnwidth]{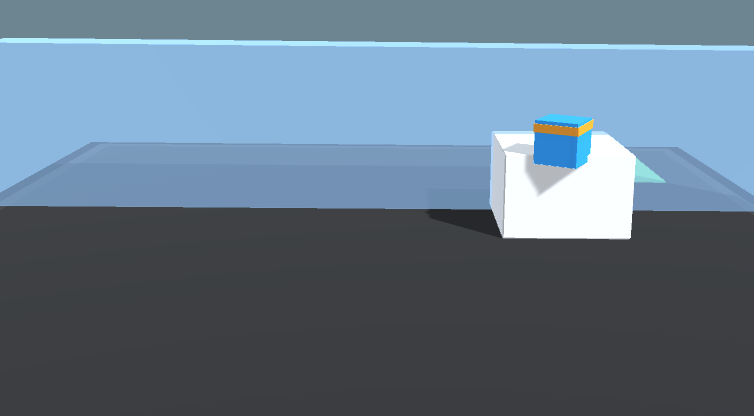}}
   }
    \caption{``Inertial'' problem of auto-curriculum gradually grows the difficulty at fixed intervals. The performance of the auto-curriculum (orange curve) significantly drops when navigation requires jumping over the box first, but the learning inertial prevents it from adapting to the new task. Note that the testing curve is evaluated on the ultimate task unless otherwise stated.}
    \label{fig:inertial}
\end{figure*}

\subsection{A Simple Interactive Curriculum Framework}
\label{sec:algorithm}
Curriculum reinforcement learning is an adaptation strategy to improve RL training by ordering a set of related tasks to be learned~\cite{bengio2009curriculum}. The most natural ordering is gradually increasing the task difficulty with an automatic curriculum. However, as shown in Figure~\ref{fig:inertial-train}, the auto-curriculum quickly mastered skills when walls were low but failed to adapt when a dramatic change of skill was required (Figure~\ref{fig:inertial-wall}), leading to a degradation of performance on the ultimate task (Figure~\ref{fig:inertial-test}). The reason is that the agent must use a box to navigate a high wall in contrast to low-wall scenarios, where additional steps to locate the box will be penalized. 

\vspace{0.8em}
\begin{algorithm}
\SetAlgoLined
\KwResult{Agent's policy $\pi^R$}
Initialize difficulty=0\;
\While{step $\leq$ total\_step}{
    $\pi^R_{new}$ = Train\label{algo:train}($\pi^R_{old}$, difficulty)\;
    \If{step \% interval ==0}{
        difficulty=$\mathcal{H}$~\label{algo:human}($\pi^R_{new}$, difficulty)\;
    }
    $\pi^R_{old}$ = $\pi^R_{new}$ \\
}
\caption{Human-Guided Interactive Curriculum}
\label{alg:algo-pipe}
\end{algorithm}

\vspace{1em}
Our results testify to what~\cite{bengio2009curriculum} observed in their curriculum for the supervised classification task, that curriculum should be designed to focus on ``interesting'' examples. In our case, the curriculum that resided at an easy level for the first 3M steps ``overfitted'' the previous skill and prevented it from adapting. Although a comprehensive IF-ELSE rule is possible, in the real-world, where situations could be arbitrarily complex, adaptable behavior out of guidance from a human is desired. Following this spirit, we test the ability of human interactive curriculum using a simple framework (Algo~\ref{alg:algo-pipe}), where human (function $\mathcal{H}$) provides feedback by adjusting the task difficulty at a fixed interval in the training loop (i.e., after evaluating the agent's learning progress on current difficulty, user can choose to tune the task easier/harder or leave it unchanged). We show in the next Section that with this simple interactive curriculum, tasks that are originally unsolvable can be guided towards success by humans, with an additional property of better generalization.

\section{Experiments} \label{sec:experiment}
We train the agents for three competitive tasks using the training method described previously. We aim to show that a human-in-the-loop interactive curriculum can leverage human prior during adaptation which allows agents to build on past experiences. For all our experiments, we fix the interaction interval (e.g., 0, 0.1, 0.2,...,0.9 of the total steps) and allow users to inspect learning progress twice before adjusting the curriculum. The user can either choose to make it easier, harder or unchanged. Our baseline is PPO with the optimized parameters as in~\cite{juliani2018unity}. We train GridWorld, Wall-Jumper, and SparseCrawler for 50K, 5M, and 10M steps, respectively.

\begin{figure*}[!h]
    \centering
    \subfloat[GridWorld (obstacles of 5)]{\label{fig:exp:gridworld}
    \includegraphics[width = 0.73\columnwidth]{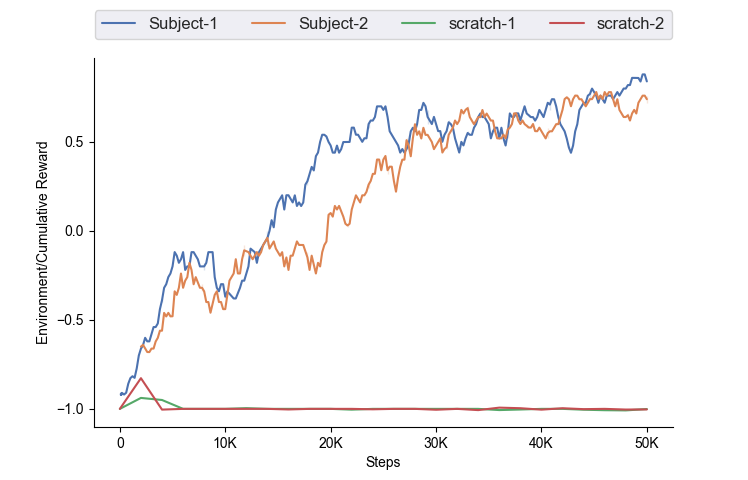}}%
    \subfloat[Wall-Jumper (height of 8)]{\label{fig:exp:jumper}
    \includegraphics[width = 0.73\columnwidth]{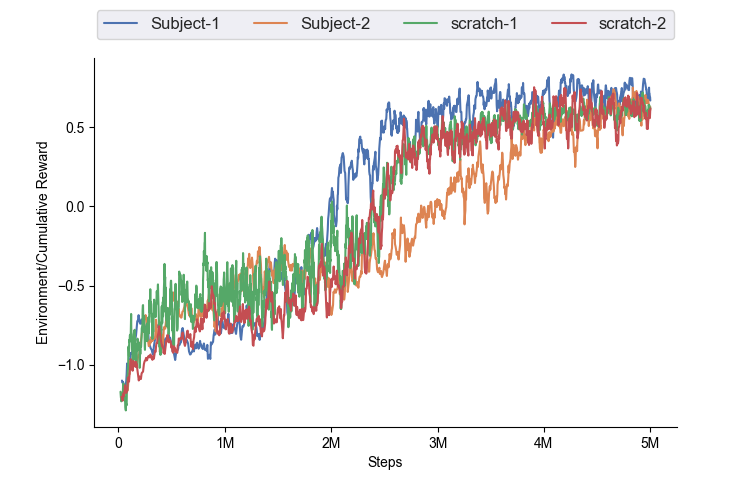}} 
   \subfloat[SparseCrawler (radius of 40)]{\label{fig:exp:crawler}
   \includegraphics[width = 0.73\columnwidth]{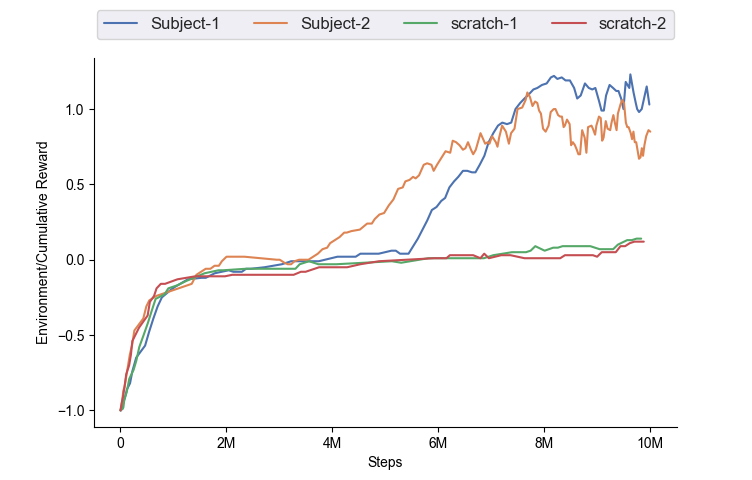}}
    \caption{\textbf{Effect of interactive curriculum evaluated on the ultimate task.}}%
    \vspace{-0.25cm}
    \label{fig:three-environment}
\end{figure*}

\begin{figure*}[!t]
    \centering
    \subfloat[GridWorld (obstacles from 1 to 5)]{\label{fig:gen:gridworld}
    \includegraphics[width = 0.73\columnwidth]{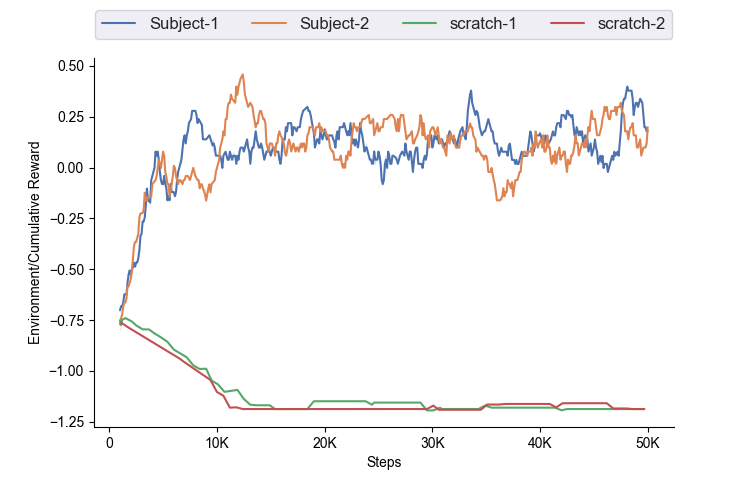}}
   \subfloat[Wall-Jumper (heights from 0 to 8)]{\label{fig:gen:jumper}.
   \includegraphics[width = 0.73\columnwidth]{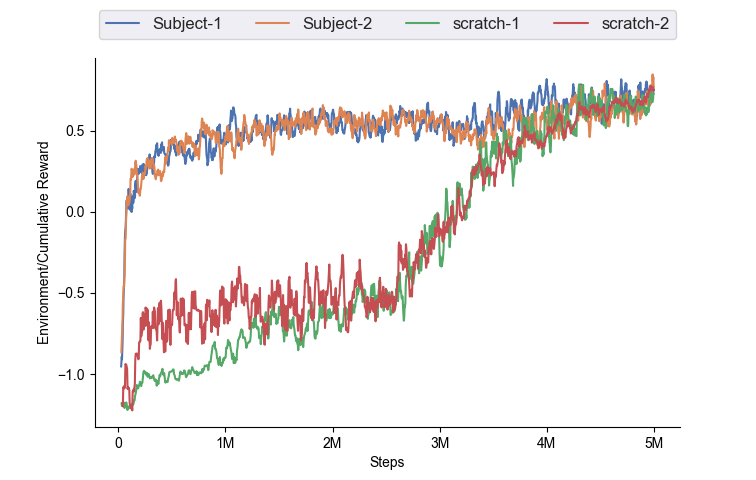}}%
   \subfloat[SparseCrawler (radius from 5 to 40)]{\label{fig:gen:crawler}
   \includegraphics[width = 0.73\columnwidth]{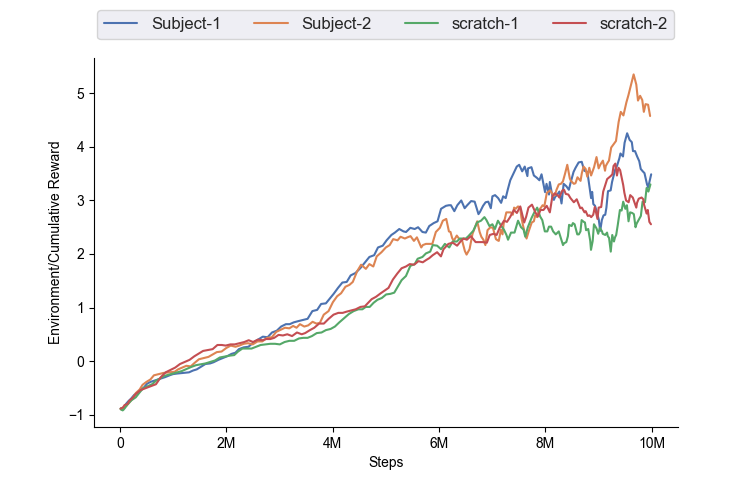}}%
    \caption{\textbf{The generalization ability of an interactive curriculum evaluated on a set of tasks. The average performance over these tasks is plotted for different time steps.}}%
    \vspace{-0.25cm}
    \label{fig:three-evaluation}
\end{figure*}

\subsection{Effect of Interactive Curriculum}
In Section~\ref{sec:interactive-platform}, we introduced three challenging tasks due to the sparsity of rewards. For example, in Figure~\ref{fig:exp:gridworld}, we observed that agents which learn from scratch (green and red curves) had little chance of success with obstacles scattered around the grid, thus failing to reinforce any desired behavior. On the other hand, users could gradually load or remove obstacles by inspecting the learning progress. Eventually, the models trained with our framework can solve GridWorld with 5 obstacles present. Inspired by this, we further tested our framework on the SparseCrawler task (Figure~\ref{fig:exp:crawler}), which requires 10M steps of training. Thanks to our parallel design (Section \ref{sec:interactive-platform}), we were able to reduce the training time from 10 to 3 hours, during which users would interact ten times. When trained with dynamically moving targets of increasing radius, we found that crawlers gradually learned to align themselves in the right direction. 

In the Wall-Jumper task (Figure~\ref{fig:exp:jumper}), we noticed a variance in performance given different users. One run (blue curve) outperformed learning from scratch with a noticeable margin, while another run (orange curve) performed less well but still converged with learning from scratch. Nevertheless, both the two trials are much better than an auto-curriculum that suffers from over-fitting, as described in Section~\ref{sec:algorithm}.

\subsection{Generalization Ability}
Over-fitting to a particular dataset is a common problem in supervised learning. Similar problems can occur in reinforcement learning when there's no or slight variation in the environment. To deal with this problem, we had considered: 1) randomness in terms of how the grid is generated; layout of blocks and jumpers; locations of crawlers and targets. 2) entropy regularization in our PPO implementation, making a strong baseline. 

We compare models trained with our framework with ones trained from scratch in three environments with a set of tasks. For example, in GridWorld, the agents were tested with the number of obstacles increasing from 1 to 5. In Wall-Jumper, the heights of the wall rise from 0 to 8 discretely during testing and in SparseCrawler, the radius of the moving target transitions from 5 to 40 with a span of 5 (Figure~\ref{fig:three-evaluation}). 
One common observation is that our model consistently outperforms learning from scratch. Secondly, there's a large gap between the curves from the curriculum learning model and learning from scratch (Figure~\ref{fig:gen:gridworld}), indicating that they ``warm up'' more quickly with easy tasks than directly jumping into the difficult task. The learning process is analogous to how human learns by building on past experiences. Interestingly, the curves eventually congregate in Wall-Jumper (Figure~\ref{fig:gen:jumper}), for both the curriculum model and scratch model. Finally, we observed that the performance of our model in SparseCrawler (Figure~\ref{fig:gen:crawler}) continually arose and reached the target with 1 to 2 more successes, as opposed to the Wall-Jumper environment. We would reset the environment in SparseCrawler only when it reaches the maximum time steps in a single round. 

When performing qualitative tests, our model solves the GridWorld with varying obstacles, whereas the learning from scratch model fails when the number of obstacles exceeds 3. For Wall-Jumper, our model can reach the goal with minimum steps, while the scratch model would inevitably use the block, necessary only for heights over 6.5. In the SparseCrawler environment, our model has a faster moving speed and more success, whereas the scratch model could only reach proximal targets.

\section{Conclusion}\label{sec:conclusion}

To learn a difficult task, humans have developed an easy-to-hard transition strategy to ameliorate the learning curve. Similarly, curriculum reinforcement learning leverages experience across many easy tasks before adapting its skills to more challenging ones. However, questions such as “what metric to use for quantifying the task difficulty” or “how should curriculum be designed” remain unanswered.

In this research, we experimented and demonstrated how human decision-making can help curriculum reinforcement learning agents make very fine-grained difficulty adjustments.
We released a multi-platform portable, interactive and parallelizable tool which features three non-trivial tasks that are challenging to solve (sparse reward, transfer between skills, and a large amount of training up to 10M steps), with varying curriculum space (discrete/continuous). We identified a phenomenon of over-fitting in auto-curriculum that leads to deteriorating performance during skill transfer with this environment. Then, We proposed a simple interactive curriculum framework facilitated by our unified user interface. The experiment shows the promise of a more explainable and generalizable curriculum transition by involving human-in-the-loop on tasks that are otherwise nontrivial to solve. We would like to explore a more efficient method for collecting users' decision-making for future work.

\section{Limitations}\label{sec:limitations}
Due to the limited user sample size and time complexity required for our environment, the project will need many resources to deploy massively. However, we believe this research can still serve as examples for human-in-the-loop reinforcement learning research.
\bibliographystyle{IEEEtran}
\bibliography{curriculum}
\end{document}